\documentclass{llncs}
\usepackage{graphicx}
\usepackage{amssymb}
\usepackage{amsmath}
\usepackage{enumitem}
\usepackage{makecell}
\usepackage{url}
\usepackage{wasysym}
\usepackage{makecell}
\usepackage{multirow}
\usepackage{wrapfig}
\usepackage{caption}
\usepackage{lipsum}
\usepackage{pifont}
\newcommand{\cmark}{\ding{51}}%
\newcommand{\xmark}{\ding{55}}%

\begin{document}
\title{Competition \textit{vs.} Concatenation in Skip Connections of Fully Convolutional Networks}
%

%

\author{Santiago Estrada\inst{1,2}\and
Sailesh Conjeti\inst{1,2}\and
Muneer Ahmad \inst{1,2}\and
Nassir Navab \inst{2}\and
Martin Reuter\inst{1,3}}

\institute{
German Center for Neurodegenerative Diseases (DZNE), Germany\and 
Computer Aided Medical Procedures, Technische Universit\"at M\"unchen, Germany \and
Department of Radiology, Harvard Medical School, Boston MA, USA
}




\maketitle              
\begin{abstract}

Increased information sharing through short and long-range \textit{skip} connections between layers in fully convolutional networks have demonstrated significant improvement in performance for semantic segmentation. In this paper, we propose Competitive Dense Fully Convolutional Networks (CDFNet) by introducing competitive maxout activations in place of na\"{i}ve feature concatenation for inducing competition amongst layers. Within CDFNet, we propose two architectural contributions, namely competitive dense block (CDB) and competitive unpooling block (CUB) to induce competition at local and global scales for short and long-range skip connections respectively. This extension is demonstrated to boost learning of specialized sub-networks targeted at segmenting specific anatomies, which in turn eases the training of complex tasks. We present the proof-of-concept on the challenging task of whole body segmentation in the publicly available VISCERAL benchmark and demonstrate improved performance over multiple learning and registration based state-of-the-art methods.

\end{abstract}
%
%
%


\section{Introduction}

Fully convolutional neural networks (F-CNNs) are being increasingly adopted for pixel/voxel-wise semantic segmentation of images in an end-to-end fashion. F-CNNs are typically constructed with a dumb-bell like architecture comprising of the encoder and decoder blocks in sequence~\cite{balanceloss}. One of the main architectural advances has been the introduction of connectivity amongst and within these blocks, which has in turn improved parameter optimization and gradient flow. Computational graph elements associated with such a connectivity can be broadly categorized into long-range and short-range connections. Long-range connections were first introduced by Ronnerberger \textit{et al.}~\cite{unet} as skip connections between the encoder and decoder blocks and were demonstrated to improve information recovery and gradient flow. Short-range connections between convolutional layers were introduced in the seminal work on residual networks by He \textit{et al.}~\cite{resnet}. This idea was taken further within the work of densely-connected neural networks~\cite{denseconnections}, wherein multiple convolutional layers were stacked in sequence along with connections that iteratively concatenate the feature maps with outputs of the previous layers. Introducing these short-range dense connections alleviate vanishing gradients, encourage feature reusability and strengthen information propagation across the network~\cite{denseconnections}. 

\begin{wrapfigure}[21]{r}{0.36\textwidth}
\vspace{-25pt}
\begin{center}
\includegraphics[width=0.36\textwidth]{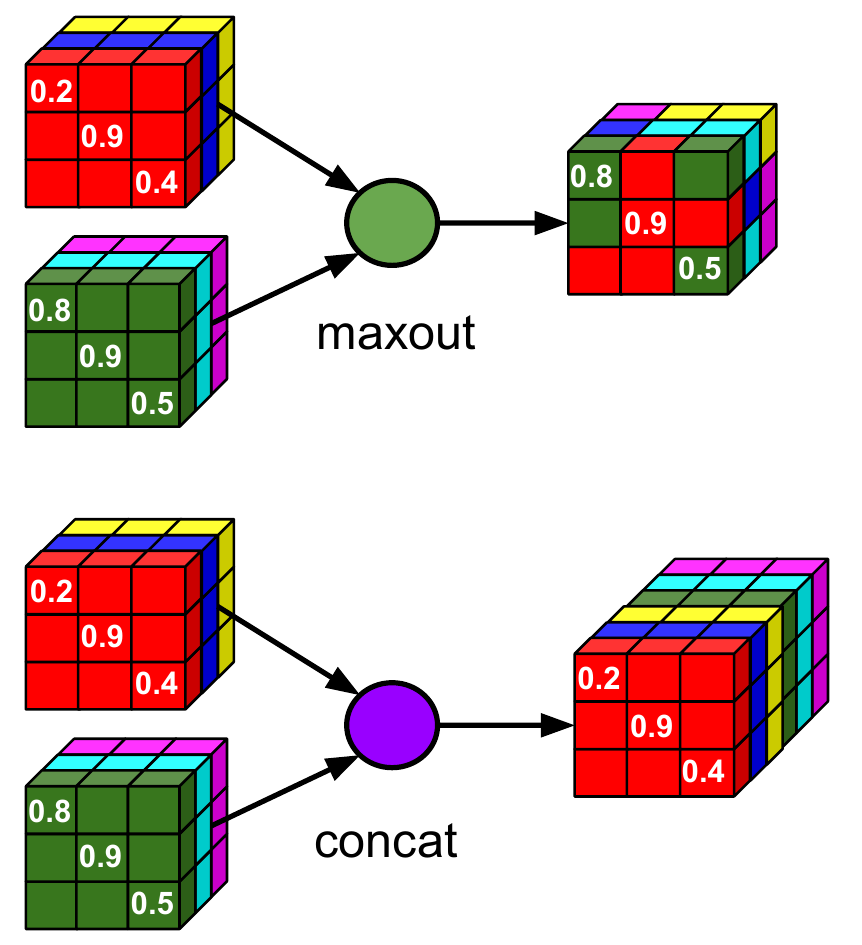}
\caption{\small{\textbf{Maxout activation}: The maxout operation computes the maximum at each spatial location across feature maps. This is a more selective fusion operation than concatenation and results in a lower dimensional feature space.}}
\label{fig:maxout}
\end{center}
\end{wrapfigure}

One commonality between design of the computational graph within the aforementioned architectures is the use of concatenation layers to aggregate information through these connections. Such a design increases the size of the output feature map along the feature channels, which in turn results in the need to learn filters with a higher number of parameters. Goodfellow \textit{et al.} introduced the idea of competitive learning through maxout activations~\cite{maxout}, which was adapted by Liao and Carneiro~\cite{competitive} for competitive pooling of multi-scale filter outputs. Both ~\cite{competitive} and~\cite{maxout} proved that the use of a maxout competitive unit boosts performance by creating a large number of dedicated sub-networks within a network that learns to target specific sub-tasks within the training task and reduces the number of parameters required. In this paper, we explore how such competitive units fare within a FCNN architecture targeted at biomedical image segmentation. We propose the Competitive Dense Fully Convolutional Network (CDFNet) by using competitive layers instead of concatenation by suitably adopting the DenseNet architecture proposed by Roy \textit{et al.} in~\cite{quicknat}. Particularly, we demonstrate that competitive units promote the formation of dedicated local sub-networks in each of the densely connected blocks within the encoder and the decoder paths. This in turn encourages sub-modularity through a network-in-network design that can learn more efficiently. Towards this, we propose two novel architectural elements targeted at introducing competition within the short- and long-range connections, as follows: 
\begin{enumerate}
\item \textbf{Local Competition}: By introducing maxout activations within the short-range skip connections of each of the densely connected convolutional layers (at the same resolution), we encourage local competition during learning of filters and the multiple convolution layers in each block prevents filter co-adaptation. 
\item \textbf{Global Competition}: We introduce a maxout activation between a long-range skip connection from the encoder and the features up-sampled from the prior lower-resolution decoder block. This promotes competition between finer feature maps with smaller receptive fields (skip connections) and coarser feature maps from the decoder path that spans much wider receptive fields encompassing higher contextual information. 
\end{enumerate}
The proof-of-concept for CDFNet is shown on the challenging task of whole-body segmentation in contrast-enhanced abdominal Magnetic Resonance Imaging (abMRI) scans as a part of the publicly available VISCERAL segmentation benchmark~\cite{visceral}.

\noindent

\section{Methodology}

\subsection{Local Competition - \textit{Competitive Dense Block}}
\subsubsection{Maxout}
The maxout is a simple feed-forward activation function that chooses the maximum value from its inputs~\cite{maxout}. Within a CNN, a maxout feature map is constructed by taking the maximum across multiple input feature maps ($\mathbf{X}$) for a particular spatial location (say $(i,j,k)$), illustrated in Fig.~\ref{fig:maxout}. Assuming $L$ inputs, denoted as $\mathbf{X} = \left \{ \mathbf{x}^{l} \right \}_{l=1}^{L}  $, with each $\mathbf{x}^{l} = \left [ x_{ijk}^{l} \right ]_{i,j,k=1}^{H,W,C}$, where $H$ is height, $W$ is width and $C$ are number of channels for a particular feature map($\mathbf{x}^{l}$). The $\text{maxout}(\mathbf{X})$ output is given by: 
\begin{equation}
\text{maxout}(\mathbf{X}) = \left [ y_{ijk} \right]_{i,j,k=1}^{H,W,C} \text{ where }y_{ijk} = \text{max} \left \{ x^{1}_{ijk},\cdots,x^{L}_{ijk} \right \}
\end{equation}

Comparing to ReLU activation that allows for division of the input space into two regions through competition with constant value of 0, the maxout activation can divide into as many regions as $L$, with each region activated by a dedicated input. Such an activation is demonstrated to better estimate exponentially complex functions, as each individual region acts as a specialized sub-module focusing on dedicated tasks and allowing for data-driven self-organization within the network during training~\cite{local_competion}. 



\subsubsection{Competitive Dense Block (CDB)} 

\begin{figure}[t!]
\centering
\begin{minipage}{\textwidth}
  \begin{minipage}[t]{0.65\textwidth}
\centering
\includegraphics[width=\textwidth]{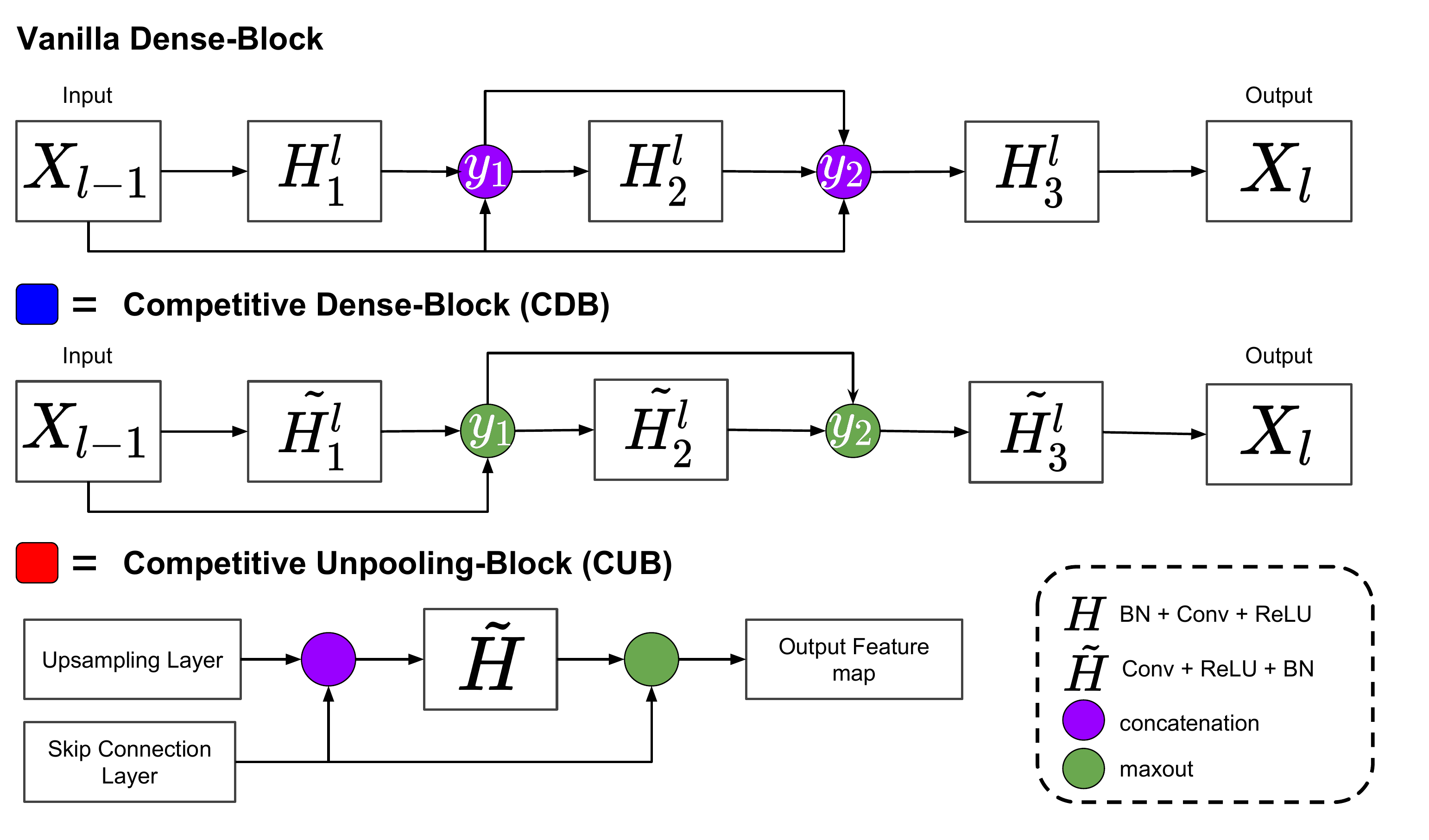}
  \end{minipage}
  \hfill
  \begin{minipage}[t]{0.35\textwidth}
\vspace{-4.4cm}
\captionof{figure}{ \scriptsize \textbf{Competitive Architectural Elements within CDFNet}: first row) Vanilla Dense Block; second row) Competitive Dense Block and third row) Competitive Unpooling Block. The red and blue squares correspond to the blocks on  Fig. \ref{fig:architecture} }
\label{fig:competitiveblocks}
    \end{minipage}
  \end{minipage}
  \vspace{-0.4cm}
\end{figure}

The dense convolutional block proposed in~\cite{denseconnections} introduces feed-forward connections from each layer to every other layer. The dense block \textit{concatenates} feature-maps of all previous layers as input to the current layer and the output of the current layer is used as input to all subsequent layers within the block (dense connections). We replace the feature map \textit{concatenations} with maxout activations to promote local competition amongst the layers. This is mathematically formulated in Eq.~\ref{eq:2} -~\ref{eq:densevsmaxout} and illustrated in Fig.~\ref{fig:competitiveblocks}. 
\vspace{-5pt}
\begin{eqnarray}
\mathbf{X}_{l} = H_{3}^{l}(\mathbf{y}_{2}) & &\mathbf{X}_{l} = \tilde{H_{3}^{l}}(\mathbf{y}_{2})\label{eq:2}\\ 
\mathbf{y}_{2} = [{H_{2}^{l}}(\mathbf{y}_{1}),\mathbf{y}_{1},\mathbf{X}_{l-1}] & &\mathbf{y}_{2} = \text{maxout} (\tilde{H_{2}^{l}}(\mathbf{y}_{1}),\mathbf{y}_{1}) \\ 
\underbrace{\mathbf{y}_{1} = [{H_{1}^{l}}(\mathbf{X}_{l-1}),\mathbf{X}_{l-1}]}_{
\text{Densely Connected Block}} & \text{        } &\underbrace{\mathbf{y}_{1} = \text{maxout} (\tilde{H_{1}^{l}}(\mathbf{X}_{l-1}),\mathbf{X}_{l-1})}_{\text{Competitive Dense Block}}\label{eq:densevsmaxout}
\end{eqnarray}

Here, $[\cdot] $ represents the concatenation operator and $\tilde{H_{j}^{l}}$ is a composite function of three consecutive operations: convolution, followed by ReLU and Batch Normalization (BN). Such a sequence of operations ensures both improved convergence while simultaneously pre-conditioning inputs to the maxout activation by ensuring an even distribution of the input points~\cite{normalisation} and an increase in the exploratory span of the created sub-networks \cite{competitive}. It must be noted that as the convolutional layers span increasing receptive fields as we traverse through the block a soft constraint is imposed to implicitly prevent filter co-adaptation.

\noindent

\subsection{Global Competition - \textit{Competitive Un-pooling block} (CUB)}

As mentioned in~\cite{unet,densenet}, the long-range skip connections between encoding and decoding paths is usually performed through the \textit{concatenation} layer. To induce competition within this layer, a na\"{i}ve solution would be to perform a maxout operation directly between the feature maps of the upsampling path and the skip connection as in the CDB design. However, we empirically observed that such architecture was unstable and resulted in loss of information. To counter this, we propose to first learn a joint feature-map (through a $1 \times 1$ convolutional layer $\tilde{H}$), which in turn competes with the features from the skip connection. Such a design (Fig.~\ref{fig:competitiveblocks}) improved feature selectivity between fine-grained with local span and coarser high-context information with much wider span coming from the up-sampling path.



\begin{figure}[t]
\centering
\includegraphics[width=\textwidth]{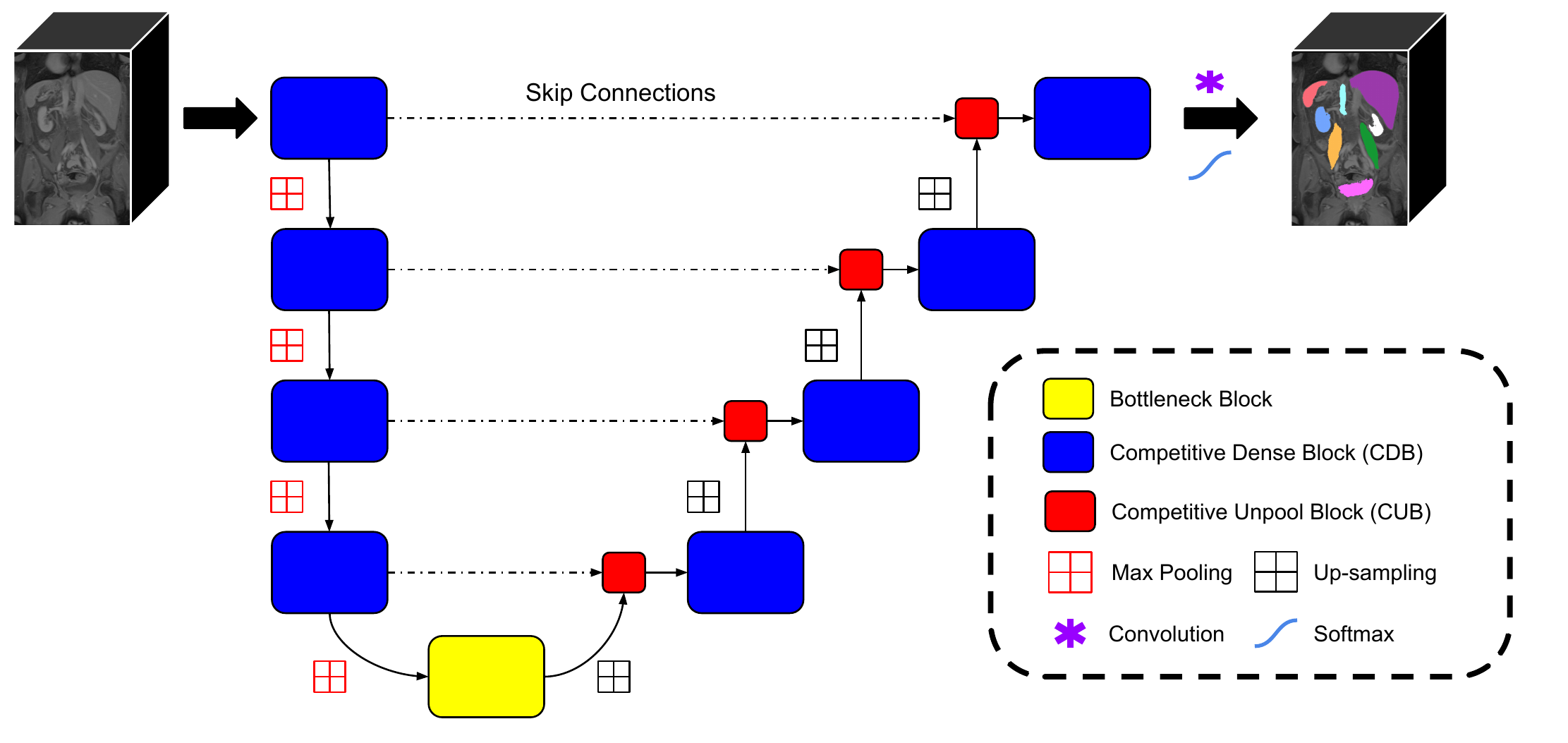}
\caption{\scriptsize \textbf{Network Architecture}: Competitive Dense Fully Convolutional Network (CDFNet), with 4 CDB on each  of the encoder and decoder path and  4 CUB between them. CDB and CUB induce local and global competition within the network.} \label{fig:architecture}
\vspace{-20pt}
\end{figure}

\subsection{Competitive Dense Fully Convolutional Network- CDFNet}

We adopt the densely connected network for semantic segmentation architecture proposed in~\cite{quicknat} and suitably introduce the CDB and CUB in place of the vanilla dense block and the unpooling layers respectively as illustrated in Fig.~\ref{fig:architecture}. In brief, the proposed CDFNet comprises of a sequence of four CDBs, constituting the encoder path (downsampling block) with four CDBs constituting the decoder path joined via a bottleneck layer. The skip connections from each of the encoder blocks feed into the CUB; that subsequently forwards features into the corresponding decoder block of the same resolution.


\section{Results and Discussion}
{\bf Dataset:} We use the abdominal T1 contrast-enhanced MR scans from the publicly available VISCERAL segmentation benchmark~\cite{visceral} for validating CDFNet. The dataset presented 13 different anatomical structures but only 10 structures were chosen for evaluation (the left out organs were annotated in less than 30\% of gold corpus volumes). The volumes were divided into patient-space splits of 15 scans for training and 5 held out for testing. Auxiliary labels available through the VISCERAL silver corpus from 70 anatomical scans were used to pre-train our models. It must be noted that the silver corpus labels were inherently noisy as they were generated by consensus fusion of the results of multiple competing algorithms~\cite{visceral}. The choice of this dataset for proof-of-concept is motivated by multiple factors 1) the task is very challenging due to potential soft organ motion (hence potential artefacts during acquisition), 2) spans a myriad of anatomies  and 3) the high degree of class imbalance increases the complexity (\textit{e.g.} liver to gallbladder has a ratio of 225:1). Moreover, the labels within the gold corpus are non-exhaustive due to potentially missing annotations in some scans. 


\noindent
{\bf Baselines and Comparative Methods:} We compare our CDFNet with state-of-the art fully convolutional networks for semantic segmentation such as densely connected network (DenseNet)~\cite{densenet}, U-Net~\cite{unet} and SD-Net~\cite{sdnet}. All the aforementioned networks were implemented maintaining consistency in the architecture \textit{i.e.} four stages of encoders and corresponding decoders feeding into the classification layer. In addition to these, methods based on multi-atlas registration (M-AR)  and label propagation from the original VISCERAL challenge (namely, M-AR via MRF and M-AR w/DOSS)~\cite{visceral} were also included for comparison.

We also test the importance of local and global competition by defining three ablative baselines: BL0: vanilla densely connected network proposed in~\cite{quicknat} (\textit{sans} any competitive blocks), BL1: network inducing local competition through CDB albeit with vanilla unpooling through concatenation and skip layers, and BL2: network inducing global competition through CUB  with vanilla dense blocks. All the aforementioned architectures were trained with a composite loss function of median frequency balanced logistic loss and Dice loss~\cite{sdnet}, together with affine data augmentation. All networks were implemented on Keras~\cite{keras} and trained until convergence using an NVIDIA Titan Xp GPU with 12 GB RAM with the following parameters: batch-size of 4, momentum set to 0.9, weight decay constant to $10^{-6}$, with an initial learning rate of 0.01 and decreased by one order every 20 epochs.


\begin{table}[t!]
\centering
\caption{Mean and standard deviation of the Dice scores for the different models and best algorithms from the VISCERAL Benchmark~\cite{visceral} on all, non-occluded and occluded organs.}
\label{dice_table}
\begin{minipage}{\textwidth}
\centering
\resizebox{0.80\textwidth}{!}{
\begin{tabular}{|c|c|c|c|c|}
\hline
{\bf Models}     & {\bf All}  & {\bf Non-occluded} & {\bf Occluded} \\ \hline
M-AR via MRF~\cite{visceral}\footnote{{\small Multi-Atlas Registration via Markov Random Field, Right Psoas Muscle excluded}}  & 0.559$\pm$0.301  & 0.777$\pm$0.120      &0.286$\pm$0.208           \\ \hline
M-AR w/DOSS~\cite{visceral}\footnote{{\small Multi-Atlas Registration w/discrete optimization and self-similarities, Occluded organs only Gallbladder}}    & -       & 0.809$\pm$0.054      & 0.494$\pm$0.238           \\ \hline
UNet~\cite{unet}          & 0.693$\pm$0.200       & 0.828$\pm$0.068      &0.491$\pm$0.146           \\ \hline
SD-Net~\cite{sdnet}        & 0.718$\pm$0.179       & 0.835$\pm$0.070      &0.543$\pm$0.138             \\ \hline
DenseNet~\cite{densenet}      & 0.731$\pm$0.184       & {\bf 0.851$\pm$0.062}     &0.550$\pm$0.153              \\ \hline
CDFNet     & {\bf 0.742$\pm$0.166} & 0.848$\pm$0.060      &{\bf 0.583$\pm$0.143} \\ \hline
\end{tabular}}
\end{minipage}
\vspace{+4pt}

\centering
\caption{Mean and standard deviation of the Dice scores for the different CDFNet baselines.}
\label{dice_tableBaselines}
\resizebox{\textwidth}{!}{
\begin{tabular}{|c|c|c|c|c|c|}
\hline
{\bf Networks} & {\bf \makecell{Local \\ Competition} } & {\bf \makecell{Global \\ Competition}} & {\bf All}  & {\bf Non-occluded} & {\bf Occluded}  \\ \hline
BL 0     &  \xmark   & \xmark    &0.731$\pm$0.184       & 0.851$\pm$0.062    &0.550$\pm$0.153  \\ \hline
BL 1  & \cmark   & \xmark   & 0.729$\pm$0.178       & 0.843$\pm$0.056      &0.559$\pm$0.152                \\ \hline
BL 2  & \xmark   & \cmark  & 0.739$\pm$0.170       & {\bf 0.852$\pm$0.061}      & 0.570$\pm$0.129            \\ \hline
CDFNet         & \cmark   & \cmark   & {\bf 0.742$\pm$0.166} & 0.848$\pm$0.060      &{\bf 0.583$\pm$0.143}              \\ \hline
\end{tabular}}
\vspace{-10pt}
\end{table}

\noindent
\textbf{Results:} To better understand the behavior of the methods towards highly varying anatomies, we categorized the target organs into non-occluded and occluded organs (organs that are most susceptible to organ motion and not clearly visible due to poor lateral resolution). Table~\ref{dice_table} presents the mean Dice scores of all organs, non-occluded organs  and occluded organs as evaluated on the held-out test data. The results of our ablative testing against local and global competition is tabulated in Table~\ref{dice_tableBaselines}. From Table~\ref{dice_table} we observe that the proposed CDFNet demonstrates the best overall Dice score in comparison to all the other comparative methods and particularly performed well in segmenting occluded organs, with a statistically significant margin ($p < 0.001$) in comparison to the closest comparative method (DenseNet), without increase in the number of parameters.

It must also be noted that all the FCNN based methods significantly out-performed M-AR based methods which is consistent with observations made in~\cite{sdnet}. From Table~\ref{dice_tableBaselines}, we infer that introducing competition simultaneously at both local and global scales improves overall performance most notably for occluded organs. Particularly, BL2 with global competition through competitive unpooling improves significantly over BL0 demonstrating that features learned through the decoders do not co-adapt with features from the skip connections.

\begin{figure}[t!]
\centering
\includegraphics[width=\textwidth]{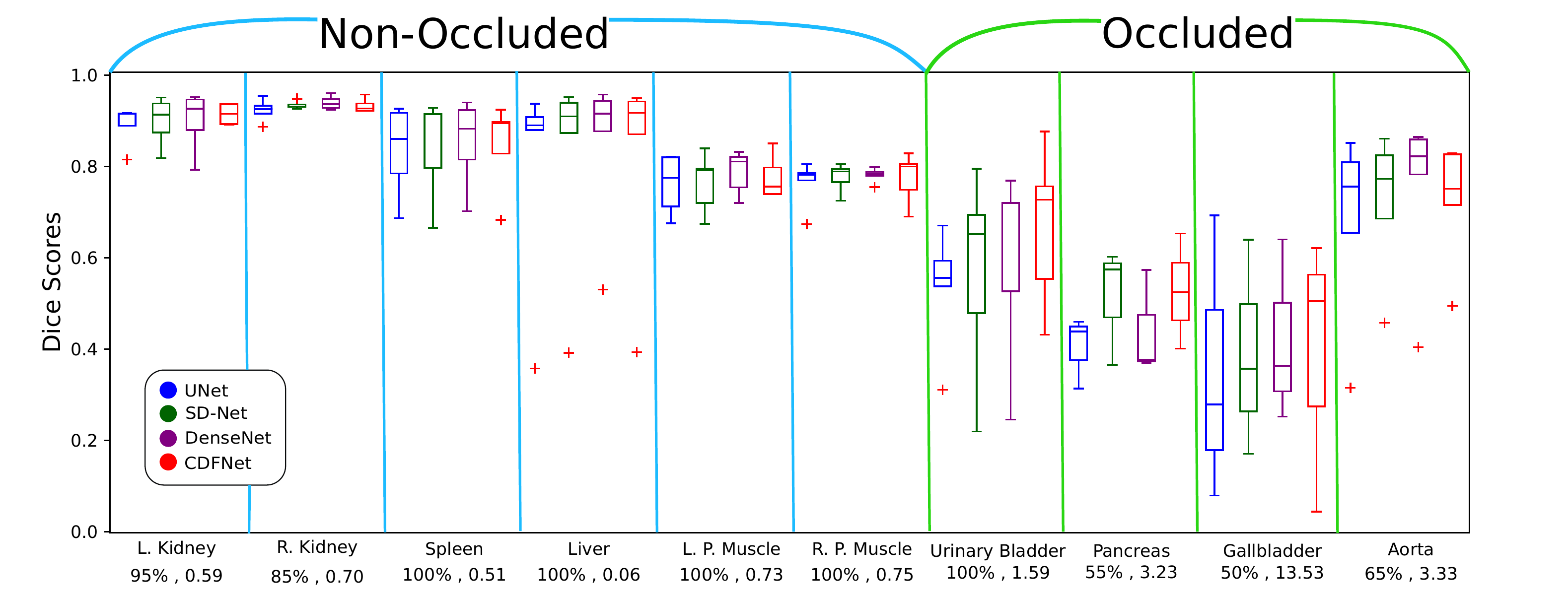}
\caption{ \scriptsize Structure-wise Dice scores boxplot comparing CDFNet {\it vs.} other FCNN architectures, Additionally the percentage of gold corpus volumes that have the particular label and degree of class imbalance are given. Left and right are indicated as L. and R. and the P. stands for Psoas Muscle.}
\label{fig:boxplot}
\vspace{-20pt}
\end{figure}

\begin{figure}
\vspace{-15pt}
\centering
\includegraphics[width=\textwidth]{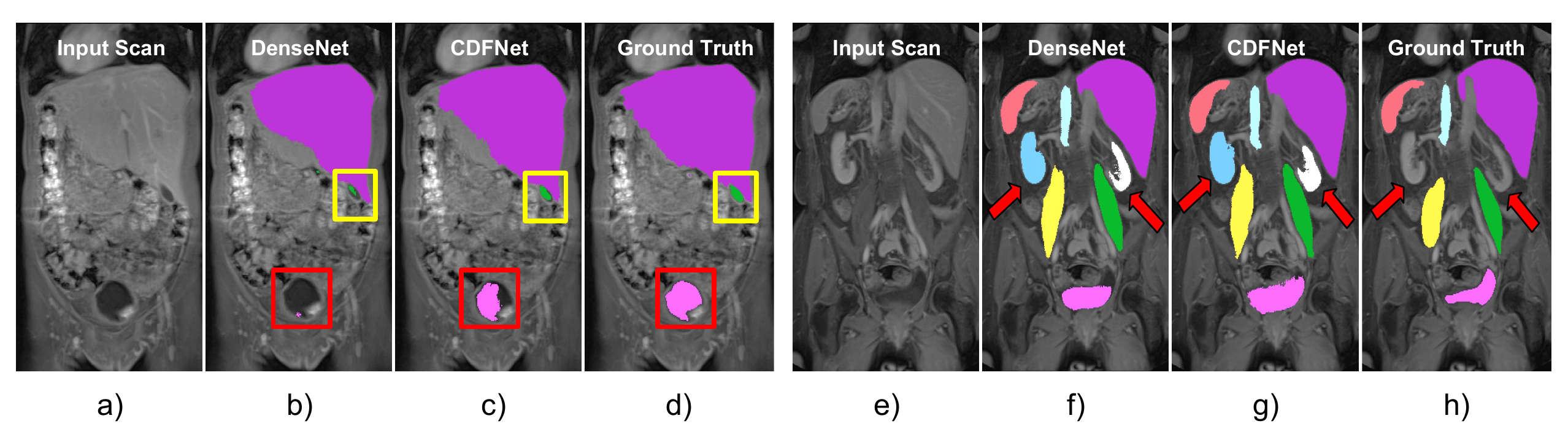}
\caption{\scriptsize Comparison of the Ground Truth {\it vs.} predictions. The red and yellow squares on (a-d)  represent the organs where the proposed method CDFNet (c) improves the segmentation over DenseNet (b). The red arrows on  left and right Kidney (e-h) show that the networks are generalizing even when they are not manually annotated on the ground truth.}
\label{fig:prediction_image}
\vspace{-30pt}
\end{figure}

Fig.~\ref{fig:boxplot} presents the structure-wise Dice scores comparing CDFNet to other FCNN architectures with additional information on the degree of class imbalance and percentage of gold corpus volumes that have the particular label. Particularly comparing  CDFNet to DenseNet, we observe that smaller and occluded organs such as gallbladder, aorta and pancreas are better recovered as competition improves network's selectivity towards fine-grained structures. We also illustrate this behavior in an unseen test scan in Fig.~\ref{fig:prediction_image}(a-d), where the networks show stark contrast in the segmentation of smaller structures, while large organs such as the liver are segmented with comparative performance. We must note that the VISCERAL Gold Corpus benchmark is not exhaustive as demonstrated in Fig.~\ref{fig:prediction_image}(h) where the left and right kidneys were not annotated despite being visible in this scan. CDFNet successfully recovers these structures as shown in Fig.~\ref{fig:prediction_image}(g).  

\section{Conclusion}
In this paper, we introduced a novel network architecture, termed Competitive Dense Fully Convolutional Network (CDFNet) that introduced competition amongst filters to improve feature selectivity within a network. CDFNet introduced competition at a local scale by substituting concatenation layers with maxout activations that prevent filter co-adaptation and reduces the overall network complexity. It also induces competition at a global scale through competitive unpooling. We evaluated our proof-of-concept on the challenging task of whole-body segmentation and clearly demonstrated that small and highly occluded structures are recovered significantly better with CDFNet over other deep learning variants that employ concatenation layers.

\bibliographystyle{splncs04}
\bibliography{mybibliography}

\end{document}